\documentclass[preprint,12pt]{elsarticle}
\usepackage{amsfonts}
\usepackage{graphics}
\usepackage[
letterpaper, top=1in, bottom=1in, left=1in, right=1in]{geometry}
\usepackage{xcolor}
\usepackage{amssymb}
\usepackage{amsthm}
\usepackage{psfrag}
\usepackage{graphicx}
\usepackage[utf8]{inputenc}
\usepackage{enumitem}
\usepackage{bm}
\usepackage{mathtools}
\usepackage{subfigure}
\usepackage{epstopdf}
\usepackage[colorlinks,linkcolor=black,anchorcolor=black,citecolor=black]{hyperref}
\usepackage{cleveref}
\usepackage{setspace}
\usepackage{amsmath,amssymb,bbm}
\usepackage[utf8]{inputenc}
\usepackage{url}
\usepackage{color}
\usepackage{caption}

\newcommand{\be}{\begin{equation}}
\newcommand{\ee}{\end{equation}}

\begin{document}

\begin{frontmatter}


\title{GenAI4UQ: A Software for Inverse Uncertainty Quantification Using Conditional Generative Models}

\author[a]{Ming Fan}
\author[b]{Zezhong Zhang}
\author[a]{Dan Lu}
\author[b]{Guannan Zhang}

\address[a]{Computational Sciences and Engineering Division, Oak Ridge National Laboratory, Oak Ridge, TN 37831.}
\address[b]{Computer Science and Mathematics Division, Oak Ridge National Laboratory, Oak Ridge, TN 37831.}

\begin{abstract}

We introduce GenAI4UQ, a software package for inverse uncertainty quantification in model calibration, parameter estimation, and ensemble forecasting in scientific applications. GenAI4UQ leverages a generative artificial intelligence (AI)-based conditional modeling framework to address the limitations of traditional inverse modeling techniques, such as Markov Chain Monte Carlo methods. By replacing computationally intensive iterative processes with a direct, learned mapping, GenAI4UQ enables efficient calibration of model input parameters and generation of output predictions directly from observations.
The software’s design allows for rapid ensemble forecasting with robust uncertainty quantification, while maintaining high computational and storage efficiency. GenAI4UQ simplifies the model training process through built-in auto-tuning of hyperparameters, making it accessible to users with varying levels of expertise. Its conditional generative framework ensures versatility, enabling applicability across a wide range of scientific domains.
At its core, GenAI4UQ transforms the paradigm of inverse modeling by providing a fast, reliable, and user-friendly solution. It empowers researchers and practitioners to quickly estimate parameter distributions and generate model predictions for new observations, facilitating efficient decision-making and advancing the state of uncertainty quantification in computational modeling.
(The code and data are available at https://github.com/patrickfan/GenAI4UQ).

\end{abstract}


\tnotetext[fn1]{This manuscript has been authored by UT-Battelle, LLC under Contract No. DE-AC05-00OR22725 with the US Department of Energy (DOE). The United States Government retains and the publisher, by accepting the article for publication, acknowledges that the United States Government retains a non-exclusive, paid-up, irrevocable, worldwide license to publish or reproduce the published form of this manuscript, or allow others to do so, for United States Government purposes. The Department of Energy will provide public access to these results of federally sponsored research in accordance with the DOE Public Access Plan (https://www.energy.gov/doe-public-access-plan).}
\end{frontmatter}

\section{Introduction}
Inverse modeling serves as a critical approach in scientific research, enabling researchers to extract critical insights into complex systems by estimating parameters, generating predictions, and quantifying uncertainties \cite{aster2018parameter}.
Traditional approaches, such as Markov Chain Monte Carlo (MCMC) sampling coupled with surrogate models, have been widely used to infer model input parameters \cite{hararuk2014evaluation, vrugt2016markov, shi2023some, ziehn2012capability}. However, these methods face notable drawbacks, including excessive computational demands, substantial memory requirements, and complicated sampling procedures \cite{lu2012analysis}. Such limitations hinder their scalability and applicability in high-dimensional or data-intensive problems \cite{reuschen2021efficient, zhang2020improving, zhang2024novel}.

Recent advances in artificial intelligence (AI), particularly in generative-based machine learning (ML) models, have provided new opportunities for more efficient and robust inverse modeling techniques \cite{zhao2023generative}. These models aim to learn data distributions and generate realistic samples, offering a more efficient alternative to traditional techniques. Existing generative models include Variational Autoencoders (VAEs), Generative Adversarial Networks (GANs), and normalizing flows, each leveraging different mechanisms to approximate data distributions \cite{kingma2013auto, goodfellow2014generative, kobyzev2020normalizing}. VAEs use probabilistic inference to map data into a latent space for generating samples, whereas GANs employ a generator network to transform random noise into realistic samples. Normalizing flows rely on invertible transformations to convert simple distributions, such as Gaussians, into complex ones, enabling precise density estimation. Despite their potential, these models encounter several challenges. VAEs are restricted by their predefined objective functions that constrain latent space representation, while GANs often suffer from instability and mode collapse \cite{liu2024diffusion}. Normalizing flow models demand intricate architectural designs to ensure tractable likelihood calculations, making them computationally prohibitive for complex data transformations \cite{yang2024conditional}.

In response to these challenges, score-based diffusion models have emerged as an innovative generative approach \cite{song2019generative, song2020denoising, zhao2023generative}. These models directly capture data distribution dynamics by learning the gradient of the data's probability density through neural networks. By iteratively solving a reverse stochastic differential (SDE) equation, they generate high-quality samples without requiring explicit normalization. Unlike previous generative techniques, diffusion models provide greater architectural flexibility and more stable training processes \cite{ song2020score}.
However, current diffusion model implementations still face significant computational barriers. The iterative reverse sampling process demands precise score estimation at each step, while unsupervised training requires extensive storage of forward stochastic differential equation trajectories. These requirements substantially increase computational complexity and memory consumption, particularly when handling complex datasets \cite{yang2023diffusion}.

Building on these advancements, this study introduces GenAI4UQ, an efficient and user-friendly software that employs a conditional generative framework to address multiple challenges in inverse modeling. The software supports model parameter estimation, prediction variable forecasting, and robust uncertainty quantification, offering a comprehensive solution to these interconnected tasks. A key innovation of GenAI4UQ is its novel methodology for estimating the score function and generating labeled data pairs. By bypassing traditional iterative and computationally expensive techniques, the method introduces a training-free mini-batch Monte Carlo estimator to approximate the score function directly. To further enhance efficiency, the estimated score function is used to solve an ordinary differential equation (ODE), producing labeled data for model training. A fully connected neural network (FCNN) is then trained in a supervised learning framework with a simple mean squared error (MSE) loss function. Auto-hyperparameter tuning is seamlessly integrated into the training process, simplifying optimization, enhancing overall model performance, and making the software accessible to users of all expertise levels. This efficient workflow significantly reduces computational complexity and provides a practical pathway to solving inverse modeling problems more effectively.
The GenAI4UQ software offers several key advantages:

\vspace{-0.2cm}
\begin{itemize}[leftmargin=15pt]
\item \textbf{Efficient parameter estimation and forecasting}: The method generates output predictions and calibrates the model input parameters with high computational efficiency, significantly reducing the time and resources required. 
\item \textbf{Comprehensive Uncertainty Quantification}: Using its generative framework, the approach delivers ensemble forecasts that capture the full complexity of parameter uncertainties, providing valuable insights into predictive reliability.
\item \textbf{Improved Computational Efficiency}: The method minimizes processing time and memory usage, overcoming critical limitations of conventional inverse modeling approaches, making it suitable for a wide range of scientific problems.
\end{itemize}
By addressing critical computational challenges, GenAI4UQ opens new avenues for scientific research, such as earth system model calibration, surface hydrology, subsurface geological carbon storage, and beyond. It provides a scalable and efficient framework for tackling the challenges of parameter inference and uncertainty quantification, paving the way for more effective scientific modeling.

\section{Mathematical Foundation of GenAI4UQ}
\begin{figure}[h!]
\centering
\includegraphics[scale=0.5]{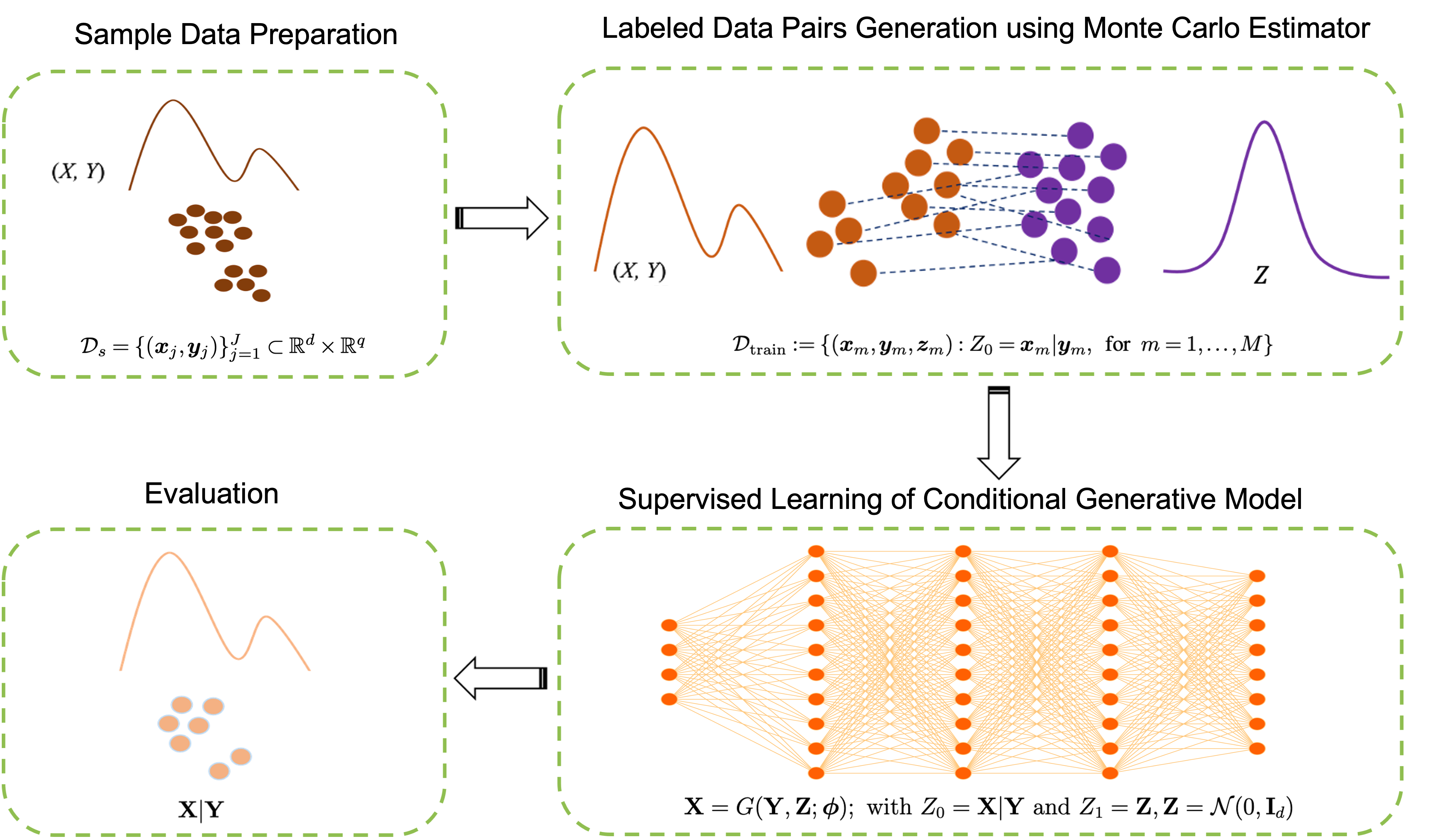}
\caption{Workflow of the proposed GenAI4UQ conditional generative framework. (a) Initial data preparation, where $X$ represents model parameters or prediction variables, and $Y$ denotes corresponding observations. (b) Labeled data pair generation using the developed Monte Carlo estimator; (c) Supervised training of a fully connected neural network to learn the conditional mapping; (d) Comprehensive model evaluation and uncertainty quantification.}
\label{fig:GenAI4UQ.png}
\end{figure}

We briefly overview the mathematical foundation of the GenAI4UQ software, and refer to our previous works \cite{YanfangSG,lu2024diffusion, fan2024novel, liu2024diffusion} for more details. Unlike traditional inverse modeling techniques, our approach efficiently generates ensemble forecasts that comprehensively capture parameter uncertainties while dramatically reducing computational complexity. The proposed GenAI4UQ conditional generative framework is illustrated in Figure \ref{fig:GenAI4UQ.png}. 
The core innovation lies in our novel score function estimation strategy. By implementing a novel Monte Carlo estimator that estimates the score function, we can generate labeled data through solving an ODE equation, effectively circumventing critical limitations of existing diffusion-based models. Through a supervised ML framework implemented via a FCNN and mean squared error optimization, we create a flexible framework for ensemble forecasts and robust uncertainty quantification. This approach represents a transformative advancement in inverse modeling, simultaneously simplifies inverse modeling processes while substantially improving computational efficiency and predictive performance across multidisciplinary scientific applications.

\subsection{Generative AI-based Conditional Framework}

This section presents the methodologies of GenAI4UQ. We develop a parametric conditional generative model $G$ leveraging a finite training dataset $\mathcal{D}_s=\{(\mathbf x_{j}, \mathbf y_{j})\}_{j=1}^{J}\subset \mathbb{R}^d\times \mathbb{R}^q$. The core objective is to train $G$ through a neural network architecture capable of generating probabilistic ensemble forecasts for the parameters of interest. Mathematically, the conditional generative model can be expressed as:

\begin{equation}\label{eq:gnmodel}
\mathbf X = G(\mathbf Y, \mathbf Z; \boldsymbol{\phi})~ \text{where}~ \mathbf Y \in \mathbb{R}^{q}, ~\mathbf Z \in \mathbb{R}^{d},
\end{equation}
where function $G$ transforms observation variables $\mathbf Y$ and a reference variable $\mathbf Z$ into target parameter variables $\mathbf X$, with $\boldsymbol{\phi}$  representing the model's parameters.

The goal is to generate ensemble forecasts for target variables based on a given observation $\mathbf y$. This is achieved by sampling $\mathbf Z$ from a standard Gaussian distribution and feeding the observation and sampled variables into the  model $G$, effectively sampling from the target conditional distribution.


\subsection{Score Function Approximation}

In probabilistic generative modeling, transforming a known distribution (such as Gaussian) to a complex target distribution involves solving a reverse-time SDE. However, the stochastic nature of this process creates non-unique mappings between initial state  $Z_0$ and the final state $Z_1$. To address this, we propose a deterministic transformation using an ODE:

\begin{equation}\label{eq:ProposedODE}
d{Z}_t = \left[ \delta(t){Z}_t - \frac{1}{2}\tau^2(t) \nabla(Z_t, t)\right] dt; \text{ with }; Z_0 = \mathbf X| \mathbf Y \text{ and } Z_1 = \mathbf Z,
\end{equation}
Where drift and diffusion coefficients are defined through alternative relationships:

\begin{equation}\label{eq:transformationcoefficients}
\delta(t) = \frac{{\rm d} \log \gamma_t}{{\rm d} t} \;\;\; \text{ and }\;\;\; \tau^2(t) = \frac{{\rm d} \rho_t^2}{{\rm d}t} - 2 \frac{{\rm d}\log \gamma_t}{{\rm d}t} \rho_t^2.
\end{equation}

To align the reference distribution with a standard Gaussian, we strategically define transformation processes:
\begin{equation}\label{eq:distributionprocesses}
\gamma_t = 1-t, \;\; \rho^2_t = t \;\; \text{ for } \;\; t \in [0,1].
\end{equation}

The score function, $S(Z_t, t)$, representing the gradient of the log probability density, is challenging to determine directly. In this work, we utilize Monte Carlo techniques to estimate the intricate score function, utilizing dataset samples to generate probabilistic representations. The estimation process involves carefully constructed weight functions that capture the underlying data distribution characteristics.
Specifically, it can be reformulated as:
\begin{equation}\label{eq:sde_exact_score}
S(Z_{t}, t) := \nabla_z \log Q_t({Z}_t) =  \int_{\mathbb{R}^d}  - \frac{Z_t- \gamma_t Z_0}{\rho^2_t} \omega_t({Z}_t,  Z_0) d Z_0,
\end{equation}

where the auxiliary weight function $\omega_t({Z}_t,  Z_0)$ is determined by
\begin{equation}\label{eq:sde_weight}
\omega_t({Z}_t,  Z_0)=\omega_t({Z}_t,  [ \mathbf X| \mathbf Y]) :=  \frac{ \displaystyle Q_{t|0}\left({Z}_t | [ \mathbf X| \mathbf Y] \right) {p( \mathbf Y|\mathbf X) p(\mathbf X)} }{\displaystyle \int_{\mathbb{R}^d} Q_{t|0}\left({Z}_t | [ \mathbf X'| \mathbf Y] \right) {p( \mathbf Y|\mathbf X') p(\mathbf X')} d  \mathbf X'| \mathbf Y},
\end{equation}
herein $p(\mathbf X)$ and $p(\mathbf Y|\mathbf X)$ represent the probability density functions of the target variable and the corresponding likelihood function.

Leveraging initial state $Z_0$ samples, the mini-batch Monte Carlo estimator is employed to approximate the score function's integrals and expectations, providing an effective estimation of the underlying probabilistic gradient. Based on the reverse-time ODE, the dataset samples $\mathcal{D}_s=\{(\mathbf x_{j}, \mathbf y_{j})\}_{j=1}^{J}$ follow the target distribution $Q_{0}(Z_0)$, originating from $Q_1(Z_1) = \mathcal{N}(0, \mathbf{I}_d)$. Therefore, $S(Z_{t}, t)$ can be approximated as:
\begin{equation}\label{eq:sde_scoreMC}
S(Z_t, t) \approx \bar{S}(Z_t, t) :=  \sum_{n=1}^{N} - \frac{Z_t - \alpha_t [\mathbf x_{j_n}|\mathbf y] }{\beta^2_t} \bar{\omega}_t({Z_t}, [\mathbf x_{j_n}|\mathbf y]). 
\end{equation}
This estimation utilizes a mini-batch $\{(\mathbf x_{j_n}, \mathbf y_{j_n})\}_{n=1}^N$ extracted from the dataset $\mathcal{D}_s$ with a batch size $N \le J$. The weight $\omega_t({Z_t},[\mathbf x_{j_n}|\mathbf y])$ is computed via:
\begin{equation}\label{eq:sde_scoreweight_app}
\omega_t({Z_t}, [ \mathbf x_{j_n}| \mathbf y]) \approx  \bar{\omega}_t({Z_t}, [ \mathbf x_{j_n}| \mathbf y]) := \frac{Q_{t|0}(Z_t | \mathbf x_{j_n}) p( \mathbf y|\mathbf x_{j_n}) p(\mathbf x_{j_n}) }{\sum_{n'=1}^{N} Q_{t|0}(Z_t|   \mathbf x_{j_{n'}}) p( \mathbf y|\mathbf x_{j_{n'}}) p(\mathbf x_{j_{n'}})} ,
\end{equation}
where $Q_{t|0}(Z_t |   \mathbf x_{j_n})$ represents the Gaussian distribution. The weight $\omega_t({Z_t},  Z_0)$ is fundamentally determined by the normalized probability density values $\{Q_{t|0}(Z_t| \mathbf x_{j_n})\}_{n=1}^N$.

\subsection{Labeled Data Pair Generation and Conditional Mapping Learning}

To learn the conditional generative model in the supervised learning framework, we need to generate labeled data pairs. First, we randomly sample $M$ reference variable samples $\mathbf Z = {z_1, \ldots, z_M}$ by drawing from a standard Gaussian distribution. Subsequently, each sample undergoes an inverse transformation via ODE integration from $t = 1$ to $t=0$ to derive the state $Z_{0}= \mathbf x_m |\mathbf y_m$, utilizing the proposed score function estimation method and the reference dataset. This process allows us to compile a generated labeled dataset that captures the relationship between initial states, observations, and reference variables.
The resulting data pairs are collected to generate the training dataset:  

\begin{equation}\label{eq:ode_label}
        \mathcal{D}_{\rm train} := \{ (\mathbf x_m, \mathbf y_m, \mathbf z_m): Z_0= \mathbf x_m |\mathbf y_m, \; \text{ for }\; m = 1, \ldots, M\}.
\end{equation}
where $Z_0$ represents the transformed initial state conditioned on the observation.

This methodical approach enables precise mapping between the input variables and the target states, facilitating subsequent model training and generative modeling. Subsequently, the generative model defined in Eq.~\eqref{eq:gnmodel} undergoes supervised training using MSE loss. Upon training completion, the model $G$ enables ensemble forecasting of target variables and uncertainty quantification by processing given observations alongside Gaussian-distributed samples.

The proposed approach offers significant computational advantages through independent labeled data generation, enabling full parallelization and dramatically reducing dataset preparation overhead. Unlike conventional MCMC methods—which often require hundreds of thousands to millions of model evaluations for new observations—our trained conditional generative model can generate ensemble forecasts in seconds, providing rapid and effective uncertainty quantification. 

\section{The GenAI4UQ Software's Features and Design Principles}

This section outlines the design and features of the GenAI4UQ software, a comprehensive ML platform designed for model calibration, uncertainty quantification, and predictive forecasting. The software integrates advanced hyperparameter optimization frameworks, overfitting prevention mechanisms, and an adaptive software architecture, making it easy to use even without prior ML knowledge. Users only need to provide their own data to leverage the software's powerful capabilities.  Key innovations include automated hyperparameter tuning using the Ray Tune framework, dynamic resource allocation for parallelized computation, and comprehensive early stopping mechanisms to prevent overfitting. The modular software architecture streamlines the workflow, allowing users to optimize models and evaluate results with minimal effort. 

\subsection{Auto Hyperparameter Tuning Framework}

Effective hyperparameter optimization is crucial for developing robust ML models, particularly in complex neural network architectures. Our research implements an advanced auto hyperparameter tuning approach utilizing the Ray Tune framework, a sophisticated scalable platform within the Ray distributed computing ecosystem. This innovative framework enables efficient parallel training of ML models across diverse computational environments, providing a robust methodology for systematically exploring and optimizing model hyperparameters \cite{liaw2018tune}.

We employed two complementary hyperparameter search strategies to determine the optimal configuration for our FCNN: grid search and random search \cite{claesen2015hyperparameter}. Grid search systematically explores hyperparameter configurations by combining predefined sets of parameter values, creating a comprehensive but computationally intensive mapping of potential model settings. Random search offers a more efficient alternative by independently sampling parameter values from a uniform distribution, particularly advantageous in high-dimensional spaces where certain hyperparameters demonstrate varying impacts on model performance. Our hyperparameter tuning strategy strategically applies these search methods to different model parameters. Grid search is employed for discrete, structural parameters such as the number of nodes, number of layers in the fully connected neural network, and batch size. Conversely, random search is utilized for continuous parameters like learning rate and dropout rate, which require more flexible exploration strategies.

Furthermore, the developed software demonstrates remarkable computational adaptability by automatically detecting and leveraging available computational resources. The framework intelligently identifies the training environment, preferentially utilizing GPU acceleration when available. In GPU-absent environments, the system seamlessly transitions to multi-core CPU processing, enabling simultaneous parallel trials to expedite hyperparameter exploration.

While the default configuration includes 10 trials, the framework provides users with extensive customization options. Researchers can dynamically adjust the number of trials, search space, and exploration strategies to suit specific research requirements. The default parameters, carefully selected to balance performance and efficiency, include a learning rate sampled from a logarithmic uniform distribution between $10^{-4}$ and $10^{-2}$, the number of neurons per layer chosen from [32, 64, 128], and one or two hidden layers. Additional settings include a dropout rate sampled uniformly between 0.01 and 0.3, batch sizes of either 32 or 64, and a maximum of 1000 epochs for training. These defaults provide a strong starting point, yielding generally excellent results in most scenarios. However, users retain full flexibility to define their own hyperparameter search configurations to address domain-specific challenges effectively.

\subsection{Overfitting Prevention Mechanisms}

We present a comprehensive approach to data preprocessing and model development, designed to ensure robust evaluation and maximize model effectiveness. The input dataset is strategically partitioned into training and testing subsets to support a reliable assessment framework. Using our developed conditional generative model, we aim to capture the underlying data distribution comprehensively through the training dataset. It is important to note that our method can generate an unlimited number of samples for training the conditional generative model $G$. These generated training data pairs follow the statistical distribution of the prepared sample data but are not identical to the original samples. By default, the system generates 20,000 training data pairs; however, users have the flexibility to customize this value according to their specific needs.

To further enhance model performance and generalization, we implement an additional data-splitting strategy during the training of the FCNN. Specifically, the generated dataset in Eq. \ref{eq:ode_label} is further subdivided into training and validation sets. This process incorporates a custom-built overfitting prevention mechanism, which systematically regulates training dynamics. By mitigating the risk of overfitting, this approach ensures that the model generalizes effectively to unseen test data, maintaining high predictive accuracy across diverse samples.

Mitigating overfitting represents a critical challenge in developing robust ML models. Our research addresses this challenge through a multi-dimensional early stopping strategy designed to ensure optimal model generalization and prevent performance degradation. The core of our overfitting prevention mechanism involves a dynamic validation loss tracking algorithm. By continuously monitoring the model's performance, the system saves checkpoints when the average validation loss demonstrates a statistically significant improvement beyond a predefined minimum delta threshold. 

Another critical component of the strategy is a comprehensive generalization gap metric that quantifying the performance difference between training and validation datasets to identify potential overfitting. Mathematically calculated as the difference between average validation and training losses, an ideal generalization gap remains minimal. This metric carefully evaluates the divergence between training and validation performance, enabling the mechanism to detect early signs of overfitting. When potential overfitting is identified, the system can trigger appropriate interventions, such as checkpoint preservation or training termination.

The last aspect of our approach is the trend detection mechanism that analyzes the contextual relationship between training and validation loss trajectories. This algorithm implements an adaptive analysis capable of detecting complex overfitting patterns. Specifically, the system interrupts training when it observes a consistent decrease in training loss concurrent with an increase in validation loss, persisting beyond a predefined patience threshold.

By implementing this comprehensive strategy, our model training process strikes an optimal balance between thorough exploration of the model's learning potential and prudent management of computational resources. The approach represents a sophisticated mechanism for monitoring and controlling the learning process, ultimately contributing to the development of more robust and generalizable ML models.

\subsection{Modular Software Architecture for Automated Model Development}

\begin{figure}[h!]
\centering
\includegraphics[scale=0.5]{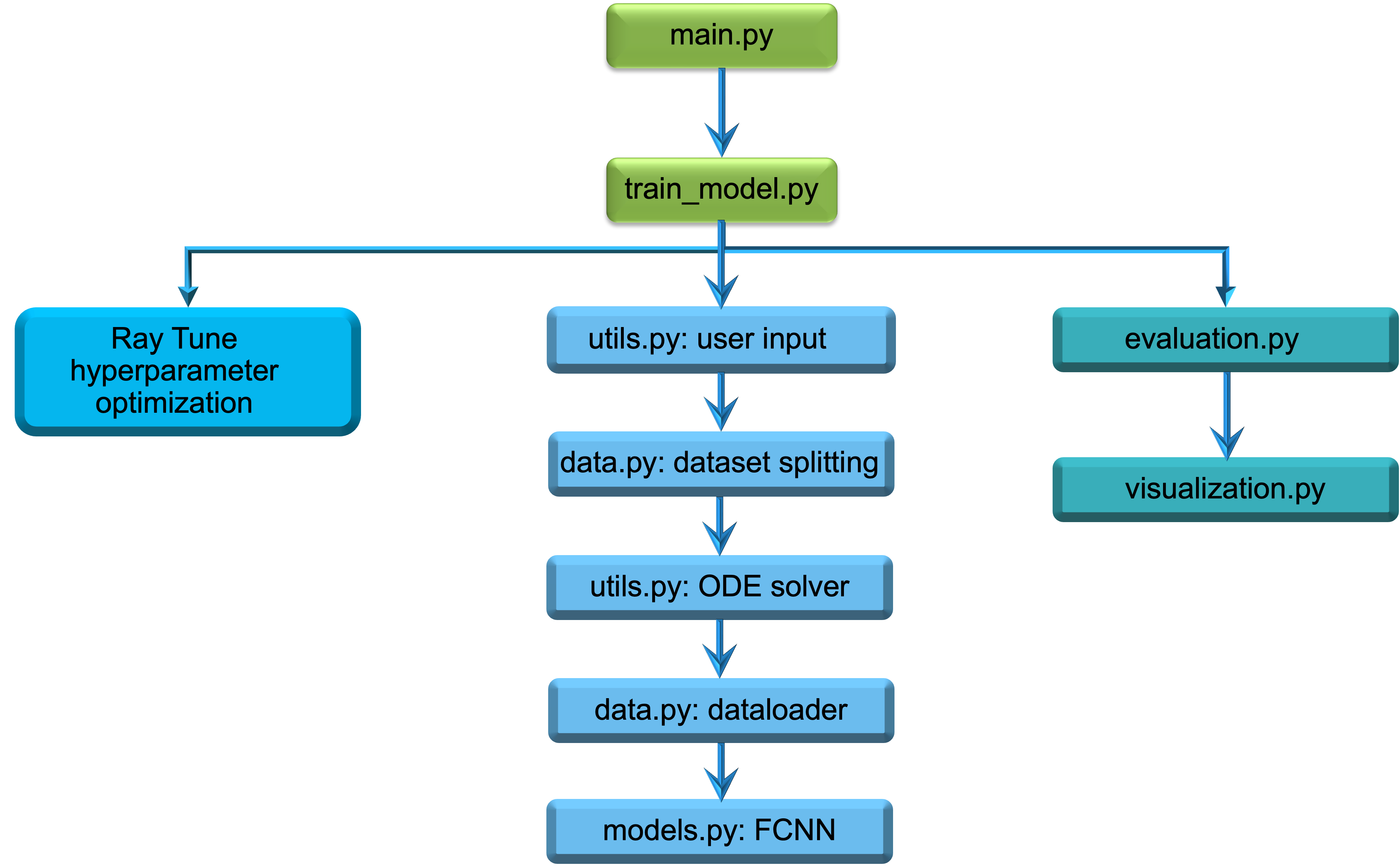}
\caption{Flow chart for software architecture and component interactions: it shows the interconnections between different modules and the sequential/parallel processing steps in the GenAI4UQ software pipeline.}
\label{fig:software_archi}
\end{figure}

The proposed software architecture is illustrated in Figure \ref{fig:software_archi}. The central control script, main.py, serves as the entry point, initiating the entire workflow that integrates multiple specialized modules for model training, hyperparameter optimization, and comprehensive evaluation. The train\_model.py module acts as a critical hub, simultaneously engaging multiple components to streamline the model development process. It first interfaces with utils.py to capture user-defined configurations, allowing dynamic specification of data locations and hyperparameter search ranges, thereby providing substantial flexibility in experimental setup. Concurrently, data.py performs critical dataset preprocessing, implementing stratified splitting techniques to generate robust training and validation datasets. A novel contribution lies in the implementation of a training-free mini-batch Monte Carlo estimator within utils.py to solve ODE, which generates labeled training data pairs from the user defined datasets and that can represent the original dataset distribution. The module leverages Ray Tune for sophisticated hyperparameter optimization, systematically exploring the predefined parameter space to identify optimal model configurations. The neural network architecture, implemented in models.py, employs a FCNN design, capable of capturing complex nonlinear relationships within the dataset. Following model training, evaluation.py conducts rigorous performance assessment, interfacing with visualization.py to generate results analysis that facilitate deeper insights into model behavior, performance metrics, and underlying data characteristics.  {\it This design empowers users with minimal ML expertise to perform sophisticated scientific computing tasks. They only need to provide their own datasets in utils.py and can leverage the software's capabilities for model calibration, uncertainty quantification, and efficient predictive forecasting.} The code can be found in the repository: https://github.com/patrickfan/GenAI4UQ.

\section{Demonstration of GenAI4UQ in Diverse Scientific Scenarios}

In this section, we demonstrate the versatility and robustness of the proposed GenAI4UQ methodology through three diverse case studies: (1) Bimodal Function Calibration, (2) Calibration of the Earth System Model-Land Model at the Missouri Ozark AmeriFlux Site, and (3) High-Dimensional Target Variable Forecasts in Geological Carbon Storage.

For each example, we detail the model training performance and evaluation of target variable estimations with uncertainty quantification, demonstrating how the methodology adapts to distinct scientific challenges while maintaining computational efficiency and prediction accuracy.

\subsection{Example 1: Bimodal Function Calibration}

We first apply GenAI4UQ to a simple yet illustrative one-dimensional bimodal problem. The forward model \( g(X) \) is defined as:
\begin{equation} 
g(X) = X^2, 
\end{equation} 
where \( X \) follows a uniform prior distribution \( U([-2, 2]) \) over the interval \([-2, 2]\). The observation variable \( Y \) is characterized by:
\begin{equation} 
Y = g(X) + \varepsilon, 
\end{equation} 
with measurement noise \( \varepsilon \) drawn from a Gaussian distribution \( \mathcal{N}(0, \sigma^2) \) characterized by \( \sigma = 0.01 \). 

\subsubsection{Data Preparation and Model Training}
The initial prior dataset \( \mathcal{D}_{\text{prior}} \) comprises 10,000 samples of \( X \) randomly drawn from the uniform distribution, paired with corresponding \( Y \) values generated through the forward model. To generate labeled data \( \mathcal{D}_{\text{label}} \), the reverse-time ODE in Eq. \ref{eq:ProposedODE} is solved to produce 20,000 samples. The labeled dataset is used to train the FCNN employing our auto hyperparameter tuning framework, which optimally configures parameters such as the learning rate, number of layers, and dropout rate. The optimal hyperparameter configuration is automatically saved for reproducibility.

\begin{figure}[h]
\centering
\includegraphics[scale=0.55]{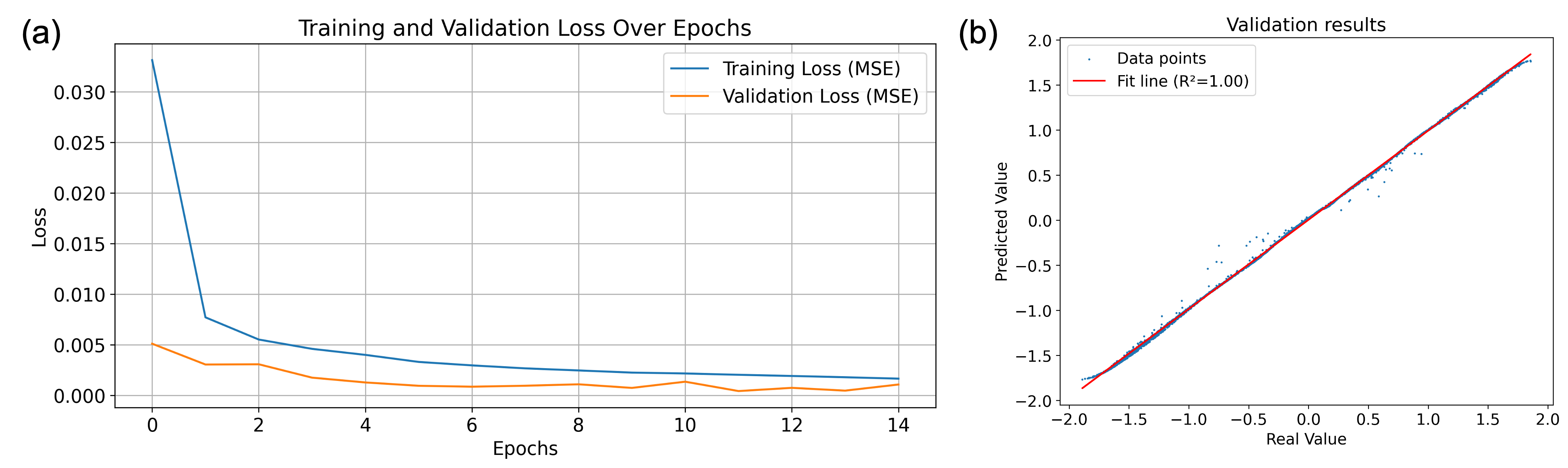}
\caption{Training performance for the bimodal case. (a) Training and validation loss curves over epochs, demonstrating stable convergence without overfitting. (b) Validation results showing the linear relationship between true and predicted values, with an \(R^2\) value close to 1, indicating the model's high accuracy in capturing the data distribution.}
\label{fig:Bimodel_sta}
\end{figure}

Figure \ref{fig:Bimodel_sta} demonstrates the training performance. In Figure \ref{fig:Bimodel_sta}a, the convergence of training and validation losses indicates effective learning without overfitting. Figure \ref{fig:Bimodel_sta}b, shows the linear relationship between true and predicted values during validation, with an $R^2$ score close to 1, illustrating the model's ability to accurately capture the underlying data distribution.

\subsubsection{Evaluation with Test Dataset}
\begin{figure}[h]
\centering
\includegraphics[scale=0.5]{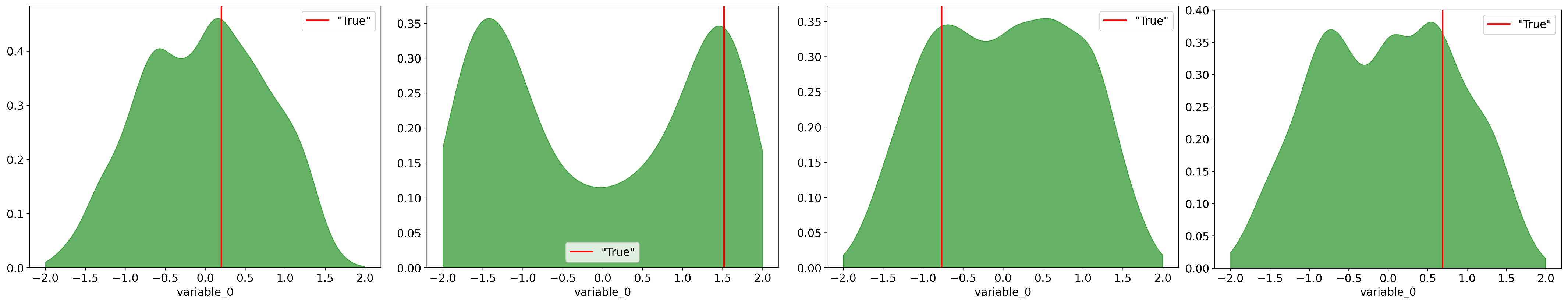}
\caption{Posterior distribution estimation for input parameter \(X\) in the bimodal case, evaluated on four randomly selected test samples. The red lines indicate the true values.}
\label{fig:Bimodel_pred}
\end{figure}

Following neural network training, we evaluate the model's performance by evaluating it on a held-out test dataset that remains completely unseen during both the labeled data generation and FCNN training processes. Figure \ref{fig:Bimodel_pred} illustrates the posterior distribution estimates for four randomly selected test cases. The results exhibit the characteristic bimodal distribution by sampling 2,000 standard Gaussian random variables through the trained generative model. 
The accuracy of the predictions is evidenced by the high density of true values lying well within the high-probability regions of the estimated posterior, particularly near the peaks of the bimodal distributions. This outcome validates the high-fidelity inference capability of our GenAI4UQ software.

\subsection{Example 2: Calibration of the Earth System Model-Land Model}

The Energy Exascale Earth System Model (E3SM) Land Model (ELM) plays a critical role in understanding ecosystem responses to climate change and in developing strategies for mitigation and adaptation \cite{golaz2019doe, leung2020introduction, lu2019efficient}. This example highlights the effectiveness of the GenAI4UQ approach in calibrating key parameters of ELM and quantifying their uncertainty using observed latent heat flux (LH) measurements from the Missouri Ozark AmeriFlux forest site. Even with a limited dataset, our method demonstrates robust parameter estimation and uncertainty quantification.

\subsubsection{Data Preparation and Model Training}

To calibrate ELM parameters, we use annual average LH measurements from 2006 to 2010 as observational data, which are  provided by Gu et al., \cite{gu2016testing}. Based on sensitivity analysis \cite{lu2024diffusion, ricciuto2018impact, xia2015joint}, eight parameters were identified for calibration. These include the rooting distribution depth factor (variable\_0), specific leaf area at the canopy top (variable\_1), fraction of leaf nitrogen in RuBisCO (variable\_2), fine root carbon-to-nitrogen ratio (variable\_3), fine root-to-leaf allocation ratio (variable\_4), base rate of maintenance respiration (variable\_5), critical day length for autumn senescence onset (variable\_6), and growing degree days required for spring leaf-out (variable\_7). Each parameter has a biologically meaningful range defined by previous studies, encompassing the diversity of environmental conditions represented in ELM \cite{lu2019efficient}.

Given observed LH data \( y \), the objective is to estimate the posterior distribution \( p(X | Y = y) \) for these eight parameters \( X \). Despite the limited dataset size  (1,000 samples), the generative framework in GenAI4UQ facilitates robust inference. Using the data, we construct a labeled dataset \( \mathcal{D}_{\text{label}} \) by solving the reverse-time ODE in Eq. \ref{eq:ProposedODE}. The FCNN generator \( G \) is trained on this dataset to learn the mapping between observations and ELM input parameter distributions. Once trained, \( G \) is used to generate 2,000 posterior samples of \( X \) from standard Gaussian random variables, effectively approximating the posterior distribution.

\begin{figure}[h]
\centering
\includegraphics[scale=0.55]{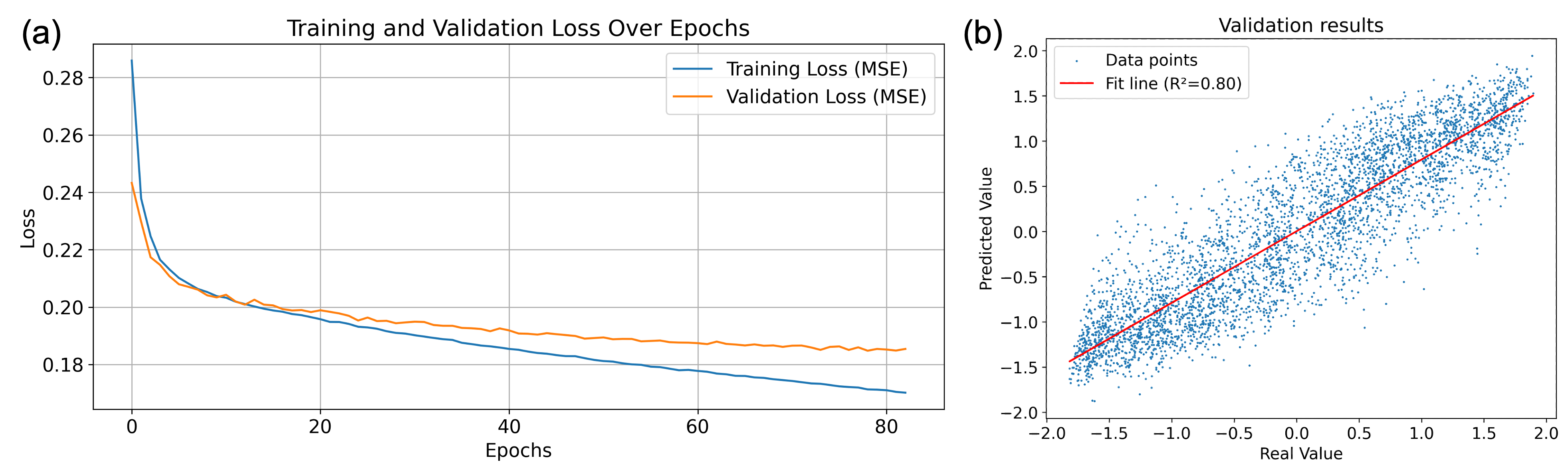}
\caption{Training performance for the ELM parameter calibration case. (a) Training and validation loss curves over epochs. (b) Validation results with an \(R^2\) analysis.}
\label{fig:ELM_sta}
\end{figure}

Figure \ref{fig:ELM_sta} demonstrates the training performance. Figure \ref{fig:ELM_sta}a shows that training and validation losses converge smoothly, indicating effective learning without overfitting.  Figure \ref{fig:ELM_sta}b illustrates the strong linear relationship between true and predicted values during validation, with an $R^2$ score close to 0.8. This high accuracy, achieved despite the limited data (1,000 sample data), demonstrates the method’s capability to learn complex parameter distributions.

\subsubsection{Evaluation with Test Dataset}

\begin{figure}[h]
\centering
\includegraphics[scale=0.5]{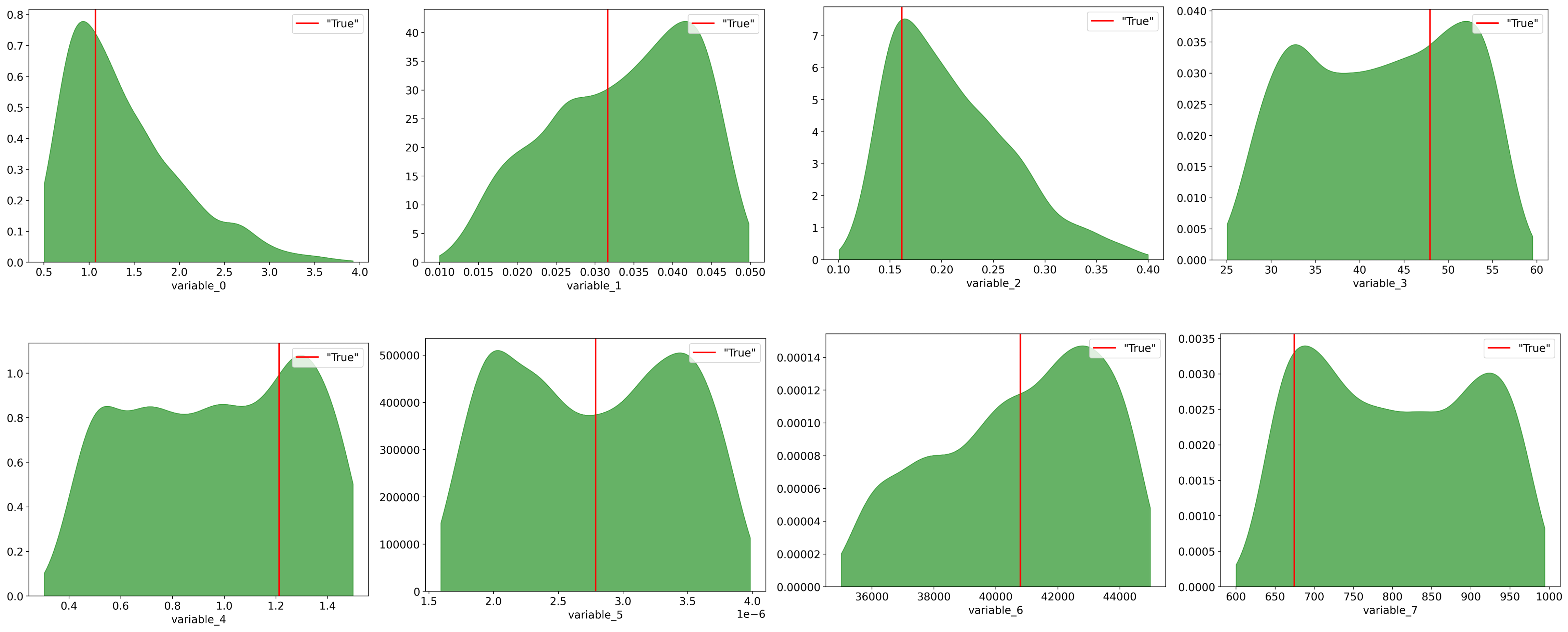}
\caption{Posterior distribution estimation for eight input parameters in the ELM model parameter calibration case, evaluated on one randomly selected test samples. The red lines indicate the true values.}
\label{fig:ELM_pred}
\end{figure}

To assess model performance, we evaluate it on a held-out test dataset. Figure \ref{fig:ELM_pred} presents the estimated posterior distributions for the eight parameters for one randomly selected test case. Each subplot shows the marginal posterior distribution for a single parameter, with the true parameter value indicated by the red line. The results confirm the accuracy of the predictions, as the true parameter values fall well within the high-probability regions of the estimated posterior distributions. This demonstrates the model’s ability to capture the underlying data distribution effectively. Additionally, the varying widths of the marginal distributions reflect the uncertainty levels associated with each parameter. For example, parameters with broader distributions indicate higher uncertainty, while narrower distributions signify greater confidence in the estimation.

These findings validate the efficacy of the GenAI4UQ framework in generating high-fidelity parameter estimates and uncertainty quantification, even when working with limited data. By capturing the intricate relationships between observations and parameter distributions, our approach offers a reliable and computationally efficient solution for ELM calibration.

\subsection{Example 3: High-Dimensional Target Variable Forecasts in Geological Carbon Storage}

Geological Carbon Storage (GCS), where CO$_2$ is captured and securely stored underground, offers an effective solution for reducing atmospheric CO$_2$ emissions \cite{fan2023deep,wang2024deep, fan2020influence}. Successful GCS deployment requires robust tools for monitoring and predictive analysis to manage risks such as pressure buildup and potential leakage, ensuring safe, permanent storage and building confidence in the technology \cite{fan2024advancing}. This process typically involves iterative calibration of reservoir models, which can be computationally intensive, especially for large-scale systems \cite{fan2024novel}. In this example, we leverage GenAI4UQ to forecast the entire pressure field distribution based on observation variables at injection wells. Despite the high-dimensional nature of the target variables, our method demonstrates remarkable performance.

\subsubsection{Data Preparation and Model Training}

The dataset used in this study is based on reservoir simulations conducted by Wen et al. (2022) \cite{wen2022u}. The numerical experiments simulate a CO$_2$ injection process into a radially symmetrical system, where supercritical CO$_2$ is injected through a vertical well with a radius of 0.1 m. The well configuration supports injection across either the full reservoir thickness or a specific depth range. Key simulation parameters include a 30-year injection duration, an injection rate between 0.2 and 2 Mt/year, and reservoir thicknesses ranging from 12.5 to 200 m. The reservoir features no-flow vertical boundaries at the top and bottom, with a vertical cell dimension of 2.08 m to account for reservoir heterogeneity. The radial extent of the reservoir is set to 100,000 m, with 200 radially discretized grid cells ensuring accurate representation of CO$_2$ plume migration and pressure buildup. The simulations generated 24 temporally distributed pressure snapshots; for this work, we focus on the final snapshot with a high-dimensional pressure distribution with dimensions of $64 \times 128$. The dataset comprises 4,500 reservoir simulations, providing diverse scenarios for training and evaluation.

To enhance computational efficiency, the high-dimensional pressure field distributions are reduced to a latent space of dimension 20 using a convolutional autoencoder with skip connections (code available in the repository). 10 observations at the injection well are extracted to serve as input features. The conditional generative model is trained on the the data pairs of the corresponding 10-dimensional observations at the injection well and corresponding 20-dimensional latent space representations. Upon the completion of training, the model $G$ can efficiently generate ensemble predictions of pressure field distributions based on injection well measurements.

\begin{figure}[h]
\centering
\includegraphics[scale=0.55]{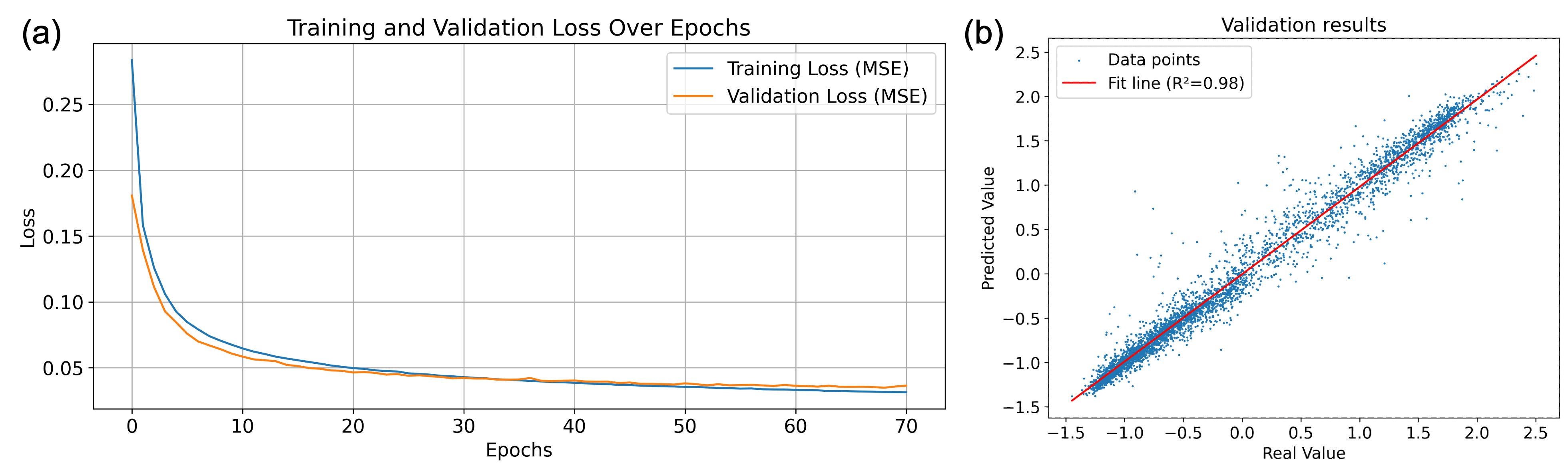}
\caption{Training performance for the High-Dimensional zParameter Estimation in Geological Carbon Storage. (a) Training and validation loss curves over epochs. (b) Validation results with an \(R^2\) analysis.}
\label{fig:GCS_sta}
\end{figure}

The training performance is illustrated in Figure \ref{fig:GCS_sta}. Figure \ref{fig:GCS_sta}a shows the convergence of training and validation loss curves, indicating effective learning without overfitting. Figure \ref{fig:GCS_sta}b highlights the strong linear relationship between true and predicted values during validation, with an $R^2$ score close to 0.98. This result demonstrates the model’s exceptional ability to learn complex parameter distributions and accurately reconstruct high-dimensional outputs.

\subsubsection{Evaluation with Test Dataset}

\begin{figure}[h!]
\centering
\includegraphics[scale=0.5]{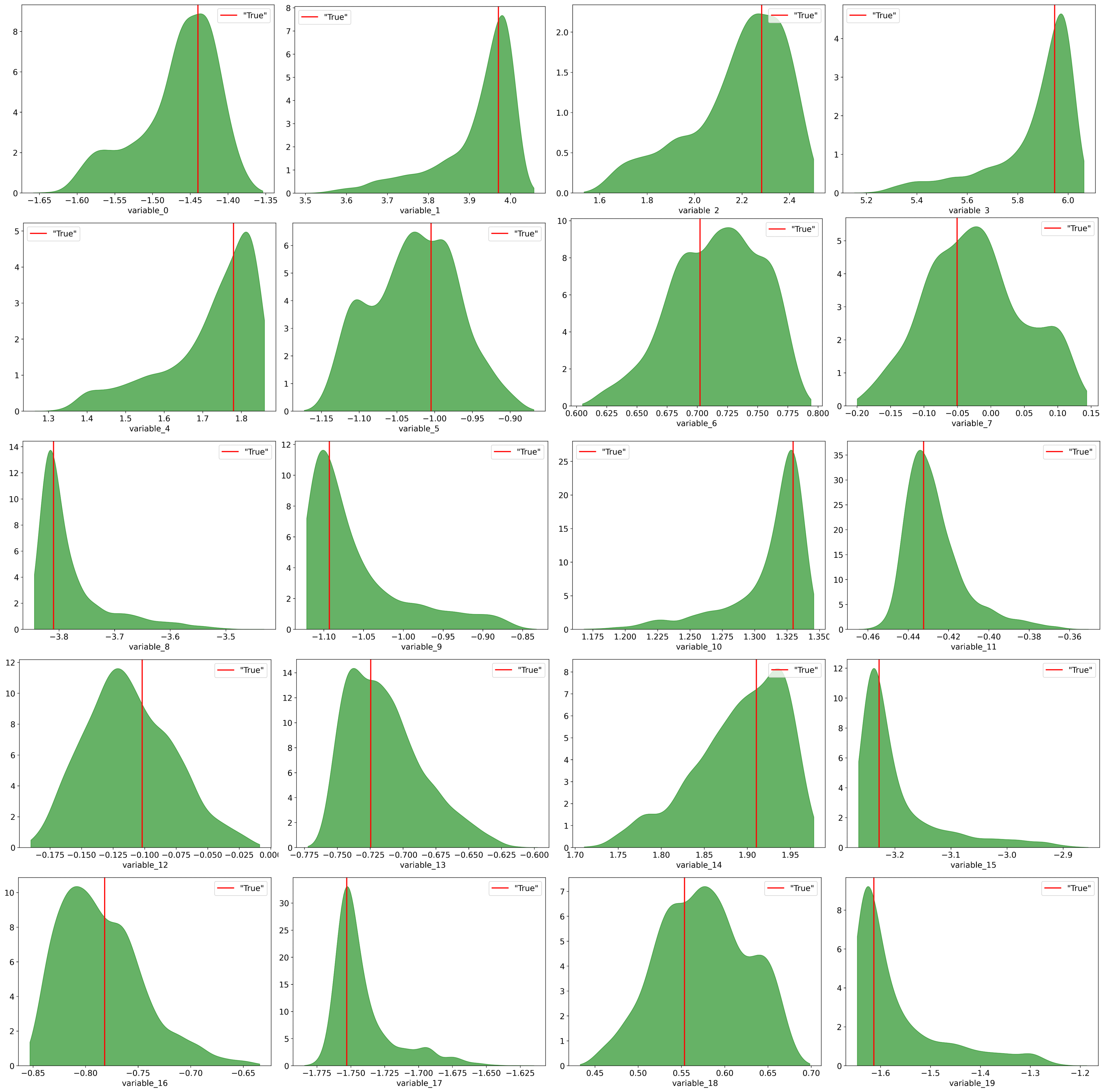}
\caption{Marginal posterior distributions for 20 latent variables in the pressure distribution forecasts for a GCS test case. The red lines mark the true parameter values.}
\label{fig:GCS_pred}
\end{figure}

The model's performance was evaluated on a held-out test dataset to ensure its generalizability and reliability. Figure \ref{fig:GCS_pred} presents the marginal posterior distributions of 20 latent variables for a randomly selected test case. Each subplot highlights the predicted probability density for an individual parameter, with the true parameter values marked by red lines. The results highlight the model’s ability to accurately capture the underlying data distribution. The true parameter values consistently fall within the high-probability regions of the estimated posterior distributions, demonstrating capability to generate accurate and uncertainty-aware predictions.  Furthermore, the varying widths of these distributions reflect parameter-specific uncertainty, offering insights into the confidence levels of individual predictions. 

\begin{figure}[h!]
\centering
\includegraphics[scale=0.7]{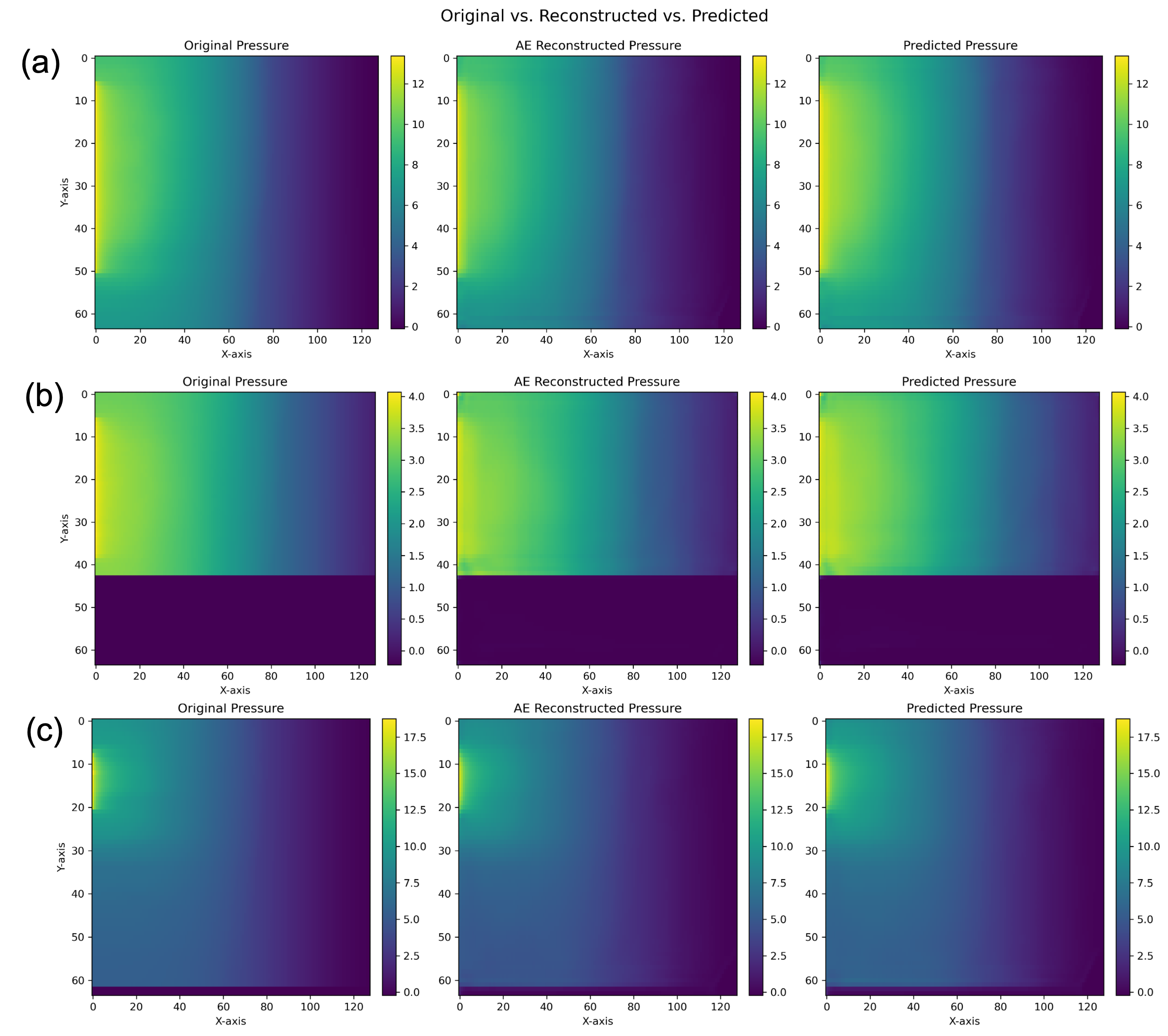}
\caption{Pressure field predictions for three test cases. The first column shows the true distributions, the second column displays the reconstructed fields using a convolutional autoencoder, and the last column illustrates the predictions from GenAI4UQ, averaged over 2,000 ensemble forecasts. }
\label{fig:GCS_dist}
\end{figure}

In addition to posterior inference, the ability of GenAI4UQ to reconstruct and predict high-dimensional pressure distributions was evaluated. Figure \ref{fig:GCS_dist} compares the true pressure distributions, the reconstructed fields from the convolutional autoencoder, and the predicted fields from the conditional generative model for three randomly selected test cases. The relatively accurate reconstructions highlight the effectiveness of dimensionality reduction, which retains essential features of the high-dimensional data while significantly reducing computational complexity. This dimensionality reduction is critical for efficient training and robust predictions, especially in scenarios with limited sample sizes and high-dimensional target variables.
The predictions (third column) closely match the ground truth pressure fields (first column), demonstrating the model’s ability to accurately forecast spatial field distributions. The high accuracy observed across all test cases indicates that the model accurately captures the spatial evolution of the pressure field under different reservoir dynamics.

These results demonstrate the potential of GenAI4UQ as an efficient and reliable tool for forecasting pressure distributions in GCS applications, enabling robust risk assessment and enhancing confidence in subsurface CO$_2$ storage technology.

\section{Conclusions}

This study introduces GenAI4UQ, a software package for tackling complex inverse uncertainty quantification challenges in model calibration, parameter estimation, and ensemble forecasting through a conditional generative AI framework. By replacing traditional inverse modeling methods with a direct, learned mapping, GenAI4UQ not only addresses computational inefficiencies but also offers a scalable and robust alternative for a wide array of scientific applications.

The effectiveness of the proposed framework is validated across three distinct examples, demonstrating its capability to accurately estimate target variable distributions and provide uncertainty-aware predictions. These case studies highlight the adaptability of GenAI4UQ to diverse problem domains, from simple bimodal problems to sophisticated environmental and geological systems. The integration of automated hyperparameter tuning and overfitting prevention mechanisms ensures optimal performance while requiring minimal user expertise in ML. Its user-friendly design, computational efficiency, and focus on inverse uncertainty quantification make it a valuable tool for researchers aiming to unlock insights from complex datasets.

\section{Data Availability Statement}
The code and data are available at https://github.com/patrickfan/GenAI4UQ.

\section{Author Contributions}
MF developed the algorithms, implemented the numerical experiments, prepared the figures, analyzed the results, and drafted the manuscript.  DL developed the algorithms, prepared the figures, and analyzed the results. GZ and ZZ developed the algorithms, prepared the figures, and analyzed the results. All authors contributed to the manuscript preparation.

\section{Acknowledgments}
This work was supported by the U.S. Department of Energy (DOE), Office of Science, Office of Advanced Scientific Computing Research, through the Applied Mathematics program under Contract ERKJ388. Additional support was provided by Dan Lu’s Early Career Project, sponsored by the Office of Biological and Environmental Research within the DOE. All research was conducted at Oak Ridge National Laboratory, operated by UT-Battelle, LLC, for the DOE under Contract DE-AC05-00OR22725.

\bibliographystyle{abbrv}
\bibliography{UQ_ref}
\end{document}